\documentclass[10pt,journal,compsoc]{IEEEtran}
\ifCLASSOPTIONcompsoc
  \usepackage[nocompress]{cite}
\else
  \usepackage{cite}
\fi

\usepackage{amsmath}\usepackage{epsfig}
\usepackage{graphicx}
\usepackage{amsmath}
\DeclareMathOperator*{\argmin}{arg\,min}
\usepackage{amssymb}
\usepackage{algorithm}
\usepackage{algpseudocode}
\usepackage{mathrsfs}
\usepackage{amssymb}
\usepackage{amsthm}\usepackage{amssymb}
\usepackage{amsthm}
\usepackage{multicol}
\usepackage{multirow}
\usepackage{subfig}
\newtheorem{theorem}{Theorem}

\theoremstyle{definition}
\newtheorem{definition}[theorem]{Definition}

\newtheorem{Problem Statement}{Problem Statement}

\ifCLASSINFOpdf
\else
\fi
\begin{document}
%
\title{Joint Concept Matching based Learning for Zero-Shot Recognition}
%
%
%
%

\author{
	\IEEEauthorblockN{Wen~Tang,~
	Ashkan~Panahi,~
	and Hamid~Krim\\
	}
	\IEEEauthorblockA{Department
		of Electrical and Computer Engineering\\
		North Carolina State University, Raleigh, NC 27606\\
		E-mail: \{wtang6, apanahi, ahk\}@ncsu.edu}
}

\IEEEtitleabstractindextext{%
\begin{abstract}
Zero-shot learning (ZSL) which aims to recognize unseen object classes by only training on seen object classes, has increasingly been of great interest in Machine Learning, and has registered with some successes. Most existing ZSL methods typically learn a projection map between the visual feature space and the semantic space and mainly suffer which is prone to a projection domain shift primarily due to a large domain gap between seen and unseen classes. In this paper, we propose a novel inductive ZSL model based on projecting both visual and semantic features into a common distinct latent space with class-specific knowledge, and on reconstructing both visual and semantic features by such a distinct common space to narrow the domain shift gap. We show that all these constraints on the latent space, class-specific knowledge, reconstruction of features and their combinations enhance the robustness against the projection domain shift problem, and improve the generalization ability to unseen object classes. Comprehensive experiments on four benchmark datasets demonstrate that our proposed method is superior to state-of-the-art algorithms.
\end{abstract}

}

\maketitle

\IEEEdisplaynontitleabstractindextext

%
\IEEEpeerreviewmaketitle

\IEEEraisesectionheading{\section{Introduction}\label{sec:introduction}}
\IEEEPARstart{I}{n} recent years, object recognition has been remarkably improved even for large-scale recognition problems, such as the ImageNet ILSVRC challenge \cite{russakovsky2015imagenet}. The latest deep neural networks (DNN) architectures \cite{sermanet2013overfeat,simonyan2014very,donahue2014decaf,he2016deep,bengio2013representation} reportedly achieve a super-human performance on the ILSVRC 1K recognition task. In spite of the existing successes, most techniques rely on the supervised training of DNNs to obtain visual representations of each category with abundant labeled training examples. In practice, however, the number of objects of different categories is not predetermined and often follows a long-tail distribution. Except for popular categories which have a large amount of training samples, most categories may have only few or no training samples. As a result, the capability of up-to-date DNN models in recognizing those scarce objects is highly limited. Additionally, it is intractable to manually collect and annotate the training samples for each object category. Simply for animal recognition problem, for example, there are about 8.7 million different animal species on earth to learn. Zero short learning (ZSL) \cite{socher2013zero,lampert2014attribute} and one/few shot learning techniques \cite{fei2006one,lake2015human,lake2015human,vinyals2016matching,ravi2016optimization} are thus proposed to overcome such limitations. In contrast to training the supervised DNNs, ZSL is to learn the capability of transferring the relevant knowledge from known objects to knowledge-less or unknown ones.

ZSL has recently received wide attention in various studies \cite{scheirer2013toward,lampert2014attribute,romera2015embarrassingly,shigeto2015ridge,changpinyo2016synthesized,zhang2016zero,chao2016empirical,akata2016label,xian2017zero}. It aims to recognize the objects with no corresponding labeled samples (unseen objects) from the knowledge obtained from the objects with a considerable number of training samples (seen objects).
In order to transfer the knowledge from seen classes to unseen ones, there is a common assumption that the side information about each class is available, such as class attributes \cite{kankuekul2012online,lampert2014attribute,frome2013devise} or word vectors \cite{frome2013devise,socher2013zero}. When both seen and unseen class names are embedded in the semantic word vector space, they are then called class prototypes \cite{fu2014transductive}. Since the distance between word vectors or class attributes is measurable, and the visual features could also be projected onto a semantic space, the idea of conventional ZSL is to learn a general mapping or relationship between the visual features and side information from labelled seen classes, and then apply it to the unseen classes. Recognizing unseen objects is performed by projecting the visual features of unseen classes into the semantic spaces by learnt mapping and assign the label by a simple nearest neighbour (NN) search, which becomes a classical classification problem.

	\begin{figure*}[!htb]
		\centering
		\includegraphics[width=0.98\textwidth]{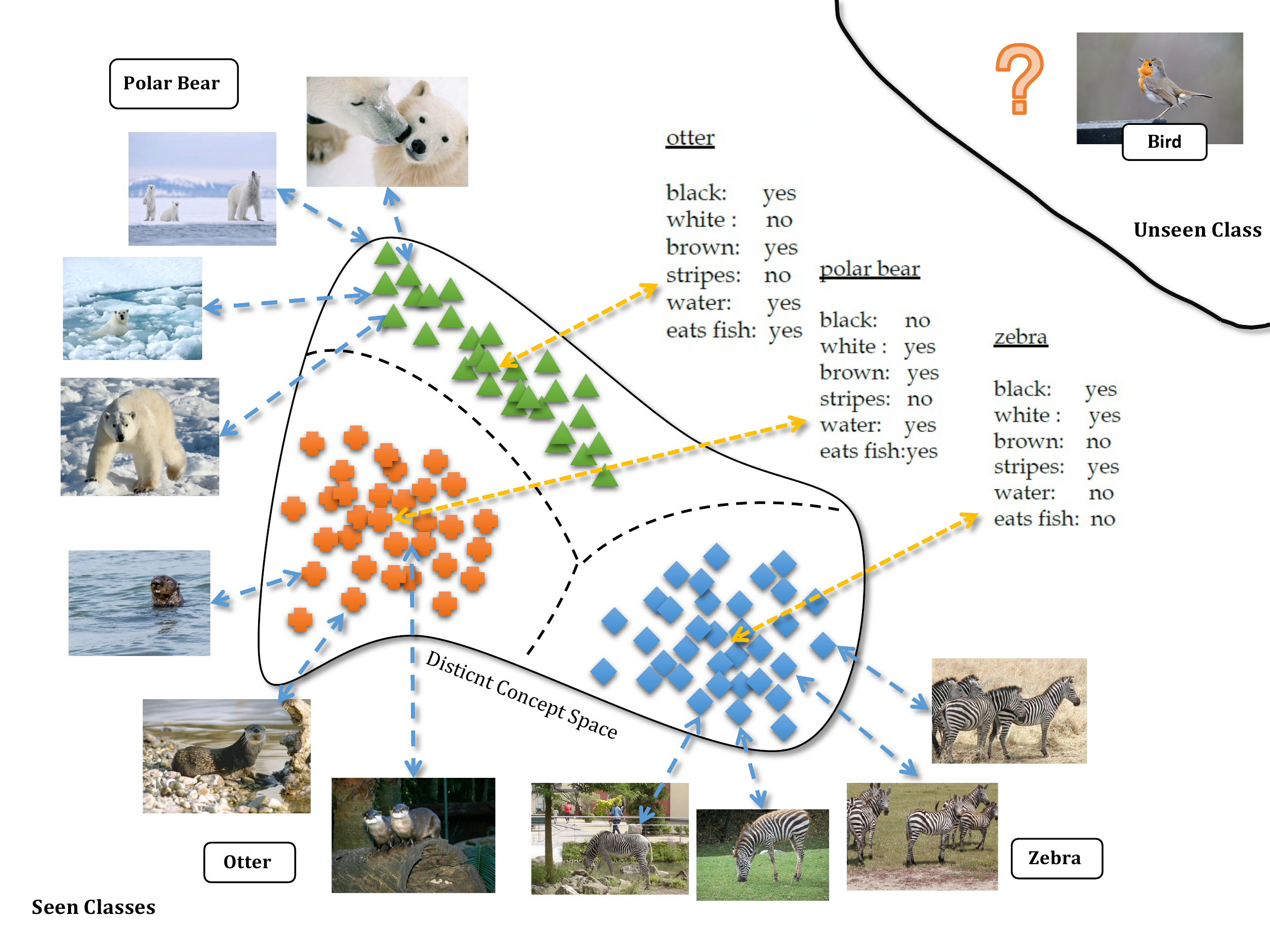}
		\caption{The Illustration of JCMSPL.}
		\label{fig:intuition}
	\end{figure*}

Most existing ZSL models \cite{lampert2014attribute,akata2015evaluation,socher2013zero,frome2013devise,reed2016learning,lei2015predicting,shigeto2015ridge,kodirov2015unsupervised,shojaee2016semi,zhang2017learning} mainly focus on the projection between visual and semantic features, when the reconstruction of the original feature is not taken into account. Although seen and unseen classes share some overlapping domain, this may cause the problem of projection domain shift. Kodirov \emph {et. al} \cite{kodirov2017semantic} recently proposed a novel method, called Semantic AutoEncoder (SAE), which includes a reconstruction constraint on the original visual features. Such a projection not only includes the mapping from the visual to semantic spaces, but also preserves the information for recovering the original visual features, mitigating the domain shift problem (this is in a sense, akin to accounting for the structure that assembles the features, i,e. the correlation among them). After that, Liu \emph{et. al} \cite{liu2018zero} restricted the projection function in SAE to be low-rank to enhance the robustness. Even though the reconstruction of visual features demonstrates its capability of alleviating the domain shift problem, \cite{kodirov2017semantic} and \cite{liu2018zero} actually also belong to the direct projection method between visual and semantic feature spaces. All above methods do not exploit the intrinsic mapping structure between the visual and semantic space. In general, when people observe an unknown object, they usually first search for similar concepts in mind, and then match it with the semantic meaning of the same concept, and conversely. These concepts have a class-to-class map on both of the visual and semantic features. As a result, in this paper, we propose a novel method named Joint Concept Matching-Space Projection Learning (JCMSPL) to mimic such human thinking behavior and take the advantage of the self-reconstruction to cope with the domain shift issue. We assume there is a common concept space incorporating distinct class concepts by introducing the class-specific information, and both visual and semantic features could be precisely projected and reconstructed by such common and distinct concepts. Hence, in such a way, the domain-invariant is introduced by the distinct concept mapping and the self-reconstruction of both visual and semantic features. In addition, such a distinct common concept space can precisely one-to-one match visual and semantic features of each class.

To have a more clear intuition, our proposed method is illustrated in Figure \ref{fig:intuition}. As shown, different categories in the visual space have many overlaps of colors and backgrounds, such as the similar color of an otter and a zebra, the same sea background of the polar bear and the otter, which all result in an non-separability issue among different classes. Similarly, in the semantic space, many objects also share common attributes. For examples, both the otter and the polar bear have in common, "brown", "water" and "eat fish" tags. Thus, a distinct concept space is adopted, where both visual and semantic features of these seen classes are projected as "separable concepts". Meanwhile, the visual and semantic features of each class are in one-to-one match, much like a human associates the visual features to words. Additionally, when considering the unseen classes (such as, birds), these have much different features in both visual and semantic spaces in comparison to the seen classes (such as, otter, polar bear and zebra), and also cause domain gap between seen and unseen classes. In order to mitigate such effects, the reconstructions of both visual and semantic features are also included, which are illustrated by the dash-line two-way arrows in Figure \ref{sec:introduction}.


Our main contributions, in this sequel, are summarized as follows:
\begin{itemize}
    \item A novel ZSL is proposed and based on the intermediate common concept space with class-specific information to better match visual and semantic features.
    \item Both visual and semantic features can be reconstructed by the common concept space to mitigate the domain-shift problem.
    \item An efficient algorithm based on Sylvester equation is developed, and is shown to achieve state-of-arts performance on four benchmark datasets for ZSL and generalized ZSL.
\end{itemize}

The balance of the paper is organized as follows: In Section \ref{sec:relatedwork}, we review the literature and background of relevance to this paper. We define the problem, formulate our novel approach and propose its algorithmic solution and recognition scheme in Section \ref{sec:formulation}. Substantiating experimental results and evaluations as well as the convergence and complexity analyses are presented in Section \ref{sec:experiments}. Finally, we provide some concluding remarks in Section \ref{sec:conclusion}.

\section{Related Work}\label{sec:relatedwork}
\subsection{Class-specific Information}
In order to improve classification accuracy, supervised learning methods \cite{jiang2013label,tang2018structured} incorporate class-specific information which makes the classifier more discriminative is expressed as a block-diagonal matrix or a simple binary label matrix. In both \cite{jiang2013label,tang2018structured}, the class-specific information makes the labels more consistent and distinct to each other. In this paper, we build on prior research to introduce a similar class-specific information matrix (a block-diagonal matrix) to support a common concept space with distinct characteristics for different classes. That is to make the projected visual and semantic features of different classes to be more discriminative.

\subsection{Projection Learning}
Various ZSL methods have recently been proposed, and can be divided into three groups primarily distinguished by their associated projection methods.
    (1) The first group employs a projection function from a visual space to a semantic space, and subsequently determines the class labels in the semantic space. The projection function can be a conventional regression model \cite{lampert2014attribute,akata2015evaluation} or a deep neural network \cite{socher2013zero,frome2013devise,reed2016learning,lei2015predicting}. This kind of projection methods are regarded as forward projection Learning.
    (2) In contrast to the first group and to alleviate the well-known hubness problem in the nearest neighbor search in high-dimensional spaces \cite{radovanovic2010hubs}, the second one is to learn the reverse projection function from the semantic space to the feature space \cite{shigeto2015ridge,kodirov2015unsupervised,shojaee2016semi,zhang2017learning}.
    (3) The third group projects both the visual and semantic features to an intermediate space \cite{changpinyo2016synthesized,lu2015unsupervised,zhang2016zerolatent}.
Our proposed JCMSPL model similarly adopts an intermediate space, which further used to generate two reconstruction constraints. By using the intermediate space, the visual features can be projected onto the semantic space, and vice versa. Therefore, JCMSPL can be viewed as a hybrid of all these three groups, which indirectly integrate both forward and reverse projections with an intermediate space for ZSL with additional and enhanced class-specific knowledge.

\subsection{Projection Domain Shift}
The aforementioned domain shift problem first noted in \cite{fu2014transductive}, has been addressed in two versions of the ZSL problems has been discussed in the literature. One is inductive ZSL \cite{kodirov2017semantic,liu2018zero}, where the projection function only relies on the seen classes, when all the unseen data is only used for testing. Another is transductive ZSL \cite{kodirov2015unsupervised,rohrbach2013transfer,zhao2018domain}, and incorporates the unlabelled unseen classes into the projection function learning to alleviate the domain shift problem. Our JCMSPL is based on the inductive ZSL and only uses the reconstruction constraints of both visual and semantic features together with class-specific information to cope with domain shift problem.

\section{Methodology}\label{sec:formulation}
\subsection{Notation}
Uppercase and lowercase bold letters respectively denote matrices and vectors throughout the paper. The transpose and inverse of matrices are respectively represented by the superscripts $\mathbf{T}$ and $-1$, as in $\mathbf{A^T}$ and $\textbf{A}^{-1}$. The identity matrix and all-zero matrix are respectively denoted by $\mathbf{I}$ and $\textbf{0}$.

	\begin{figure}[!thb]
		\centering
		\includegraphics[width=0.49\textwidth]{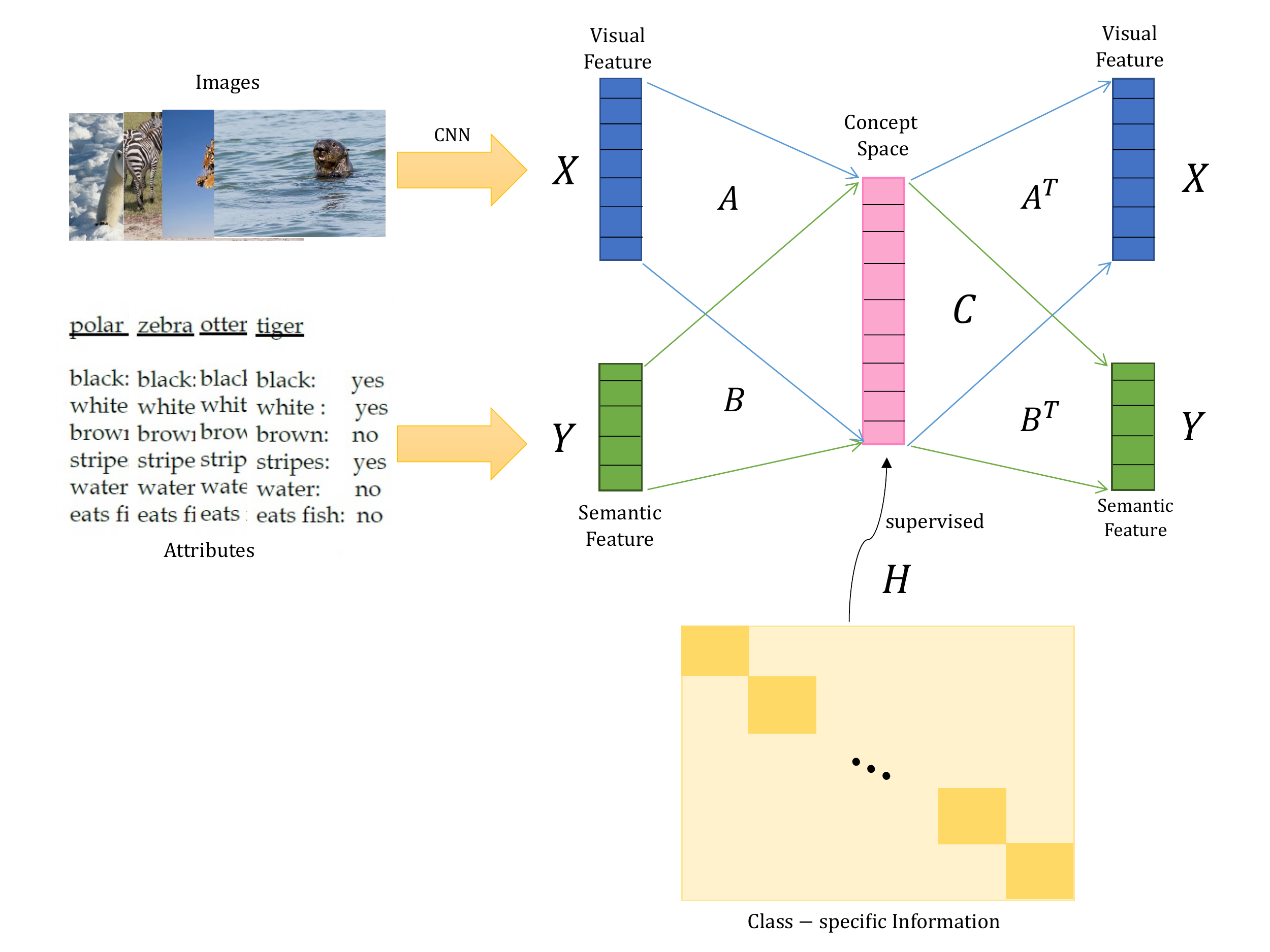}
		\caption{The framework of our purposed JCMSPL is comprised of 3 procedures. (I) A CNN is first used to extract the visual features $\mathbf{X}$, such as GoogleNet \cite{szegedy2015going}, VGG-19 \cite{simonyan2014very} and ResNet101 \cite{he2016deep}, and then project them by the function $\mathbf{A}$ into the common concept space $\mathbf{C}$. Finally, $\mathbf{A^T}$ is employed to reconstruct the $\mathbf{X}$ from concept space $\mathbf{C}$. (II) The attributes/word vector $\mathbf{Y}$ is used to embed the different classes into the semantic space. The project function $\mathbf{B}$ maps the semantic feature $\mathbf{Y}$ into the common concept space $\mathbf{C}$ as well, and $\mathbf{B^T}$ reconstruct $\mathbf{Y}$ from $\mathbf{C}$. (III) A block-diagnoal matrix $\mathbf{H}$ supervise and support the common concept space $\mathbf{C}$ to be distinct and separable, where the light-yellow part shown in the class-specific information is with all entrances of $0$ and the dark-yellow part is occupied by the elements of $1$.}
		\label{fig:framework}
	\end{figure}

\subsection{Problem Statement}
The goal of ZSL is to assign the unseen class label to the unseen samples, and both unseen labels and samples are independent from the training phase.

Let $S=\{ \mathbf{\mathbf{X_s}},\mathbf{\mathbf{Y_s}},{L_s}\}$ denote the set of seen classes with $c_s$ seen classes and $n_s$ labeled samples. And let $U=\{\mathbf{\mathbf{X_u}},\mathbf{Y_u},{L_u}\}$ denote the set of unseen classes with $c_u$ seen classes and $n_u$ labeled samples. $\mathbf{\mathbf{X_s}} \in \mathbf{R}^{m \times n_s}$ and $\mathbf{X_u} \in \mathbf{R}^{m \times n_u}$ are $m$-dimensional visual features samples in the seen and unseen sets, $\mathbf{\mathbf{Y_s}} \in \mathbf{R}^{d \times n_s}$ and $\mathbf{\mathbf{Y_u}} \in \mathbf{R}^{d \times n_u}$ are associated class-level attributes, namely semantic features. ${L_s}, {L_u}$ are respectively the corresponding label sets of seen and unseen classes. Based on the definition of ZSL, the labels of seen and unseen sets have no overlap, i.e., ${{L_s}} \cap {L_u}=\varnothing$, and ZSL aims to learn a classifier $p: \mathbf{X_u}/\mathbf{Y_u} \to {L_u}$, so as to predict the label for unseen classes, where $p$ is learned only on basis of the seen class sets $S=\{\mathbf{X_s},\mathbf{Y_s},{L_s}\}$.

\subsection{Model Formulation}
As mentioned in Section \ref{sec:introduction}, JCMSPL incorporates a distinct common concept space with class-specific information, and both visual and semantic features self-reconstructions.
Therefore, JCMSPL consists of 3 joint procedures as illustrated in Figure \ref{fig:framework}, and formulated as follows:
\begin{equation}
\label{equ:JCMSPL}
\begin{split}
    \underset{\mathbf{A},\mathbf{B},\mathbf{C}}{\min}~ &\frac{1}{2}\|\mathbf{A}\mathbf{X_s}-\mathbf{C}\|_F^2+\frac{\lambda_1}{2}\|\mathbf{B}\mathbf{Y_s}-\mathbf{C}\|_F^2+\frac{\lambda_2}{2}\|\mathbf{C}-\mathbf{H}\|_F^2\\
    &+\frac{\lambda_3}{2}\|\mathbf{X_s}-\mathbf{A^T}\mathbf{C}\|_F^2+\frac{\lambda_4}{2}\|\mathbf{Y_s}-\mathbf{B^T}\mathbf{C}\|_F^2,
\end{split}
\end{equation}
where $\mathbf{A} \in \mathbf{R}^{k\times m}$ is the projection matrix from the visual space to the common concept space $\mathbf{C} \in \mathbf{R}^{k\times n_s}$, while $\mathbf{B} \in \mathbf{R}^{k\times d}$ is the projection matrix from the semantic space to the common space. We further require the transpose matrices $\mathbf{A^T}$ and $\mathbf{B^T}$ to be respectively the reverse projection matrix from the common space $\mathbf{C}$ to the visual and semantic spaces, so that visual and semantic features can also be reconstructed by the common space. $\mathbf{H} \in \mathbf{R}^{k\times n_s}$ is a block-diagonal matrix, predefined by the class-specific information to make the common concept space distinct, and thus enhance the matching of each class more accurately. $\lambda_1,\dots,\lambda_4$ are the turning parameters.

\subsection{Algorithmic Solution}
\label{sec:optimization}
Since the objective functional in Eq. (\ref{equ:JCMSPL}) is a multi-convex problem, we may reliably update the variables by a block-coordinate descent method.

\noindent\textbf{Update A:} When $\bf{B}$ and $\bf{C}$ are fixed, $\bf{A}$ is updated by
\begin{equation}\label{equ:updateA_0}
    \mathbf{A^*}_{t+1}=\underset{\mathbf{A}}{\argmin} \frac{1}{2}\|\mathbf{A}\mathbf{X_s}-\mathbf{C}_{t}\|_F^2+\frac{\lambda_3}{2}\|\mathbf{X_s}-\mathbf{A^T}\mathbf{C}_{t}\|_F^2.
\end{equation}
As $\|\mathbf{Z}\|_F=\|\mathbf{Z^T}\|_F$, Eq.(\ref{equ:updateA_0}) can be rewritten as
\begin{equation}\label{equ:updateA_1}
    \mathbf{A^*}_{t+1}=\underset{\mathbf{A}}{\min} \frac{1}{2}\|\mathbf{A}\mathbf{X_s}-\mathbf{C}_{t}\|_F^2+\frac{\lambda_3}{2}\|\mathbf{X_s^T}-\mathbf{C}_{t}\mathbf{^T}\mathbf{A}\|_F^2.
\end{equation}
Then, taking the derivative of Eq. (\ref{equ:updateA_1}) and setting it to zero, we obtain:
\begin{equation}
    \lambda_3\mathbf{C}_t\mathbf{C}_t\mathbf{^T}\mathbf{A^*}_{t+1}+\mathbf{A^*}_{t+1}\mathbf{X_s}\mathbf{X_s^T}=(1+\lambda_3)\mathbf{C}_t\mathbf{X_s^T}.
\end{equation}
Denote, $$\mathbf{M_A}=\lambda_3\mathbf{C}_t\mathbf{C}_t\mathbf{^T},$$ $$~\mathbf{N_A}=\mathbf{X_s}\mathbf{X_s^T},$$ $$\mathbf{T_A}=(1+\lambda_3)\mathbf{C}_t\mathbf{X_s^T},$$ we have:
\begin{equation}\label{equ:updateA}
    \mathbf{M_AA^*}_{t+1}+\mathbf{A^*}_{t+1}\mathbf{N_A}=\mathbf{T_A},
\end{equation}
To solve Eq. (\ref{equ:updateA}), we use the following definition and theorems:
\begin{definition} \label{def:sylvester}
A Sylvester equation \cite{sylvester1884equation} is a matrix equation of the following form:
\begin{equation}\label{equ:sylvester}
\mathbf{RZ+ZS=T}.
\end{equation}
When $\mathbf{R}$, $\mathbf{S}$ and $\mathbf{T}$ are given, the problem is to find the possible matrices $\mathbf{Z}$ that obey this equation.
\end{definition}
\begin{theorem}
\cite{lancaster1985theory} The sufficient condition of Eq. (\ref{equ:sylvester}) to have a solution $\mathbf{Z}$ is that:

The matrix
$\begin{bmatrix}
\bf{R} &\bf{0}\\
\bf{0} &\bf{S}
\end{bmatrix}$
is similar to the matrix
$\begin{bmatrix}
\bf{R} &\bf{T}\\
\bf{0} &\bf{-S}
\end{bmatrix}$
.
\end{theorem}

\begin{theorem}\label{the:uniquesolution}
\cite{lancaster1985theory} The sufficient condition for Eq. (\ref{equ:sylvester}) to have a \textbf{unique} solution $\mathbf{Z}$ is that:

The eigenvalues $(\sigma_1^R,~\sigma_2^R,\dots,\sigma_r^R)$ of $\mathbf{R}$ and the eigenvalues $(\sigma_1^S,~\sigma_2^S,\dots,\sigma_s^S)$ of $\mathbf{S}$ satisfy $\sigma_i^R \neq - \sigma_j^S, \forall i=1,\dots,r,\forall j=1,\dots,s. $
\end{theorem}
\noindent More details and proofs of Definition 1, Theorem 1 and 2 can be found in \cite{lancaster1985theory}.

By Definition \ref{def:sylvester}, Eq. (\ref{equ:updateA}) is a Sylvester equation and easy to meet the sufficient condition in Theorem \ref{the:uniquesolution}, as ZSL is based on real image data. Eq. (\ref{equ:updateA}) is thus solved efficiently by the Bartels-Stewart algorithm \cite{bartels1972solution}, which can be implemented by a single line code: Sylvester in MATLAB\footnote{https://www.mathworks.com/help/matlab/ref/sylvester.html}.

\noindent\textbf{Update B:} When $\mathbf{A}$ and $\mathbf{C}$ are fixed, $\mathbf{B}$ is updated by
\begin{equation}\label{equ:updateB_0}
    \mathbf{B^*}_{t+1}=\underset{\mathbf{B}}{\argmin} \frac{\lambda_1}{2}\|\mathbf{B}\mathbf{Y_s}-\mathbf{C}_{t}\|_F^2+\frac{\lambda_4}{2}\|\mathbf{Y_s}-\mathbf{B^T}\mathbf{C}_{t}\|_F^2.
\end{equation}
Similarly updating $\mathbf{A}$, we have the Sylvester equation related to $\mathbf{B}$:
\begin{equation}\label{equ:updateB}
    \mathbf{{M_B}B^*}_{t+1}+\mathbf{B^*}_{t+1}\mathbf{N_B}=\mathbf{T_B},
\end{equation}
where $$\mathbf{M_B}=\lambda_4\mathbf{C}_t\mathbf{C}_t\mathbf{^T},$$ $$~\mathbf{N_B}=\lambda_1 \mathbf{Y_s}\mathbf{Y_s^T},$$ $$\mathbf{T_B}=(\lambda_1+\lambda_4)\mathbf{C}_t\mathbf{Y_s^T}.$$

\noindent\textbf{Update C:} When $\mathbf{A}$ and $\mathbf{B}$ are fixed, $\mathbf{C}$ is updated by
\begin{equation}
\label{equ:updateC_0}
\begin{split}
    \mathbf{C^*}_{t+1}=&\underset{\mathbf{C}}{\argmin}~ \frac{1}{2}\|\mathbf{A}_{t+1}\mathbf{X_s}-\mathbf{C}\|_F^2+\frac{\lambda_1}{2}\|\mathbf{B}_{t+1}\mathbf{Y_s}-\mathbf{C}\|_F^2\\
    &+\frac{\lambda_2}{2}\|\mathbf{C}-\mathbf{H}\|_F^2+\frac{\lambda_3}{2}\|\mathbf{X_s}-\mathbf{A^T}_{t+1}\mathbf{C}\|_F^2\\
    &+\frac{\lambda_4}{2}\|\mathbf{Y_s}-\mathbf{B^T}_{t+1}\mathbf{C}\|_F^2,
\end{split}
\end{equation}
Derivativing Eq. (\ref{equ:updateC_0}) and setting it to zero, yields its analytical solution as follows:
\begin{equation}\label{equ:updateC}
\begin{split}
       \mathbf{C^*_{t+1}}=&((1+\lambda_1+\lambda_2)\mathbf{I}+\lambda_3\mathbf{A}_{t+1}\mathbf{A^T}_{t+1}+\lambda_4\mathbf{B}_{t+1}\mathbf{B^T}_{t+1})^{-1}\\ &(\lambda_2\mathbf{H}+(1+\lambda_3)\mathbf{A}_{t+1}\mathbf{X_s}+(\lambda_1+\lambda_4)\mathbf{B}_{t+1}\mathbf{Y_s}).
\end{split}
\end{equation}

We follow the updating steps in each iteration of our algorithm as summarized in Algorithm 1.
\renewcommand{\algorithmicrequire}{\textbf{Input:}}
\renewcommand{\algorithmicensure}{\textbf{Output:}}
\begin{algorithm}[htb]
\label{alg:learning}
	\caption{Joint Concept Matching-Space Projection Learning}
	\begin{algorithmic}[1]
		\Require
		Visual features training set $\mathbf{X_s}$, semantic features training set $\mathbf{Y_s}$, tuning parameters $\lambda_1,\dots,\lambda_4$, and maximum iteration $t_{MAX}$.
		\Ensure
		The visual projection function $\mathbf{A}$, the semantic projection function $\mathbf{B}$ and the distinct common space $\mathbf{C}$.
		\State Initialize $\bf{A,B}$ and $\mathbf{C}$ as random matrices;
		\While {not converged \textbf{and} $t < t_{MAX}$}
		\State t=t+1;
		\State Update $\mathbf{A}_t$ by solving Eq. (\ref{equ:updateA});
		\State Update $\mathbf{B}_t$ by solving Eq. (\ref{equ:updateB});
		\State Update $\mathbf{C}_t$ by Eq. (\ref{equ:updateC});
		\EndWhile
	\end{algorithmic}
\end{algorithm}
\subsection{Zero-shot Recognition}
\label{sec:classifiy}
After we obtain the projection matrices $\mathbf{A}$ and $\mathbf{B}$, zero-shot recognition can be subsequently performed in two ways:
\begin{itemize}
    \item[(1)] With projection matrices $\mathbf{A}$ and $\mathbf{B^T}$ in hand, whenever a new test sample $\mathbf{x_{u_i}} \in \mathbf{X_u}$ is given, the associated semantic feature $\mathbf{\hat{y}_{u_i}}$ of unseen class are easily reconstructed by the visual features using the \textbf{equation}:

\begin{equation}
\mathbf{\hat{y}_{u_i}}=\mathbf{B^T}\mathbf{A}\mathbf{x_{u_i}}.
\end{equation}
The test data in the semantic space can be classified by a simple Nearest Neighbour (NN) classifier based on the distance between the estimated semantic representation $\mathbf{\hat{y}_{u_i}}$ and the prototype projections in the semantic space $\mathbf{Y_u}$. The label $\mathbf{l_{u_i}}$ for the unseen sample is assigned by,
\begin{equation}
\begin{split}
    &{l_{u_j}}=\min_{c_j}~D(\mathbf{\hat{y}_{u_j}}, \mathbf{Y_{u_{c_i}}}),\\
\end{split}
\end{equation}
where $\mathbf{Y_{u_{c_j}}}$ is the prototype attribute vector of the $c_j$-th unseen class, $D$ is an arbitrary distance function.
    \item [(2)] With projection matrices $\mathbf{A^T}$ and $\mathbf{B}$ in hand, whenever a new test sample $\mathbf{y_{u_i}} \in \mathbf{Y_u}$ is given, the associated visual feature $\mathbf{\hat{x}_{u_i}}$ of an unseen class are easily reconstructed by the semantic features thorough the following method:

\begin{equation}
\mathbf{\hat{x}_{u_i}}=\mathbf{A^T}\mathbf{B}\mathbf{y_{u_i}}.
\end{equation}
The test data in the visual space can be classified by a simple Nearest Neighbour (NN) classifier based on the distance between the estimated visual representation $\mathbf{\hat{x}_{u_i}}$ and the prototype projections in the visual space $\mathbf{X_u}$. The label $l_{u_i}$ for the unseen sample is assigned by,
\begin{equation}
\begin{split}
    &{l_{u_i}}=\min_{c_j}~D(\mathbf{\hat{x}_{u_i}}, \mathbf{X_{u_{c_j}}});\\
\end{split}
\end{equation}
where $\mathbf{X_{u_{c_j}}}$ is the $c_j$-th unseen class prototype projected in the feature space, $D$ is an arbitrary distance function.
\end{itemize}
We next validate the proposed approach of both strategies.

\section{Experiments}
\label{sec:experiments}

\subsection{Datasets and Settings}

\label{sec:datasets}
\subsubsection{Datasets}
Four benchmark datasets are used to evaluate the state-of-art methodologies along with our own.. Animals with Attributes (AwA) \cite{lampert2014attribute}, CUB-200-2011 Brids (CUB) \cite{wah2011caltech}, and SUN Attribute (SUN) \cite{patterson2014sun} are three widely used medium-scale datasets in existing ZSL works. But they are not large enough to show the capability of the original motivation of ZSL for scaling up visual recognition. Thus, the ILSVRC2012/ILSCRC2012 (ImNet) \cite{russakovsky2015imagenet} is then selected as a large-scale dataset in \cite{fu2016semi}.

AwA \cite{lampert2014attribute} consists of 30475 images of 50 animal classes with 85 associated class-level attributes, which is a coarse-grained dataset. 40 classes are used for training, while the remaining 10 classes with 6180 images are used for testing.

CUB \cite{wah2011caltech} is a fine-grained dataset with 11788 images for 200 different types of bird species which are annotated by 312 attributes. The first standard zero-shot split was introduced in \cite{akata2016label}, where 150 classes are for training and 50 classes are for testing.

SUN \cite{patterson2014sun} is also a fine-grained dataset, which includes 14340 images for 717 types of different scenes categories which are annotated by 120 attributes. Following the split in \cite{lampert2014attribute} 645 out of 717 classes are used as a training set, and the remaining 72 classes are for testing.

ImNet \cite{russakovsky2015imagenet} contains 218000 images and 1000-dimensional class-level attributes. Following the split in \cite{fu2016semi}, the 1000 classes of ILSVRC2012 are used as seen classes, when the 360 classes of ILSVRC2010 are used as unseen classes, which are not included in ILSVRC2012.

For a fair comparison against published results, we use the same above training (seen) and testing (unseen) splits for our ZSL evaluation. The summary of all those datasets are listed in the Table \ref{tab:dataset}.

\begin{table}[!htb]
		\centering
		\normalsize
		\caption{The details of four evaluated datasets. Notation: 'SS'- the semantic space; 'A'- the attribute, 'W'- the word vector; 'SS-D'- the dimension of the semantic space.}
			\resizebox{0.99\columnwidth}{!}{%
		\begin{tabular}{c c c c c}
		\hline
			\hline
			Datasets & \#Images & SS & SS-D & \# Seen/\#Unseen Classes\\
			\hline
			AwA \cite{lampert2014attribute} &  $30475$ & A &  $85$ & 40/10\\
			CUB \cite{wah2011caltech} &  $11788$ &  A  &  $312$ & 150/50\\
			SUN \cite{patterson2014sun}&  $14340$ & A &  $64$ & 645/72\\
			\hline
			ImNet \cite{russakovsky2015imagenet} &  $218000$ &  W &  $1000$ & 1000/360\\
			\hline
			\hline
		\end{tabular}%
		}
		\label{tab:dataset}%
	\end{table}%

\subsubsection{Semantic Spaces}
Generally, there are two different types of attributes. One is the attribute annotations, which is for the  medium datasets, and another one is the word vector representation, which is used for large-scale datasets. The word vector, word2vec \cite{mikolov2013distributed}, representation is obtained by training a skip-gram text model on a corpus of 4.6M Wikipedia documents.


\subsubsection{Visual Spaces}
In recent ZSL modes, all the visual features are extracted from Convolutional Neural Networks (CNNs) \cite{simonyan2014very,szegedy2015going,he2016deep} that are pre-trained by the 1K classes in ILSVRC 2012 \cite{russakovsky2015imagenet}. In our experiments, the visual features are extracted from pre-trained GoogleNet \cite{szegedy2015going}. It is worthy noting that the visual features used in most compared methods are GoogleNet features, except Table \ref{tab:starandzsl}, where a number of most ZSL models used VGG19 \cite{simonyan2014very} and ResNet101 \cite{he2016deep} features. Since the source codes of such models are not released, we can not report the results based on GoogleNet, instead the results reported in the original paper are listed in Table \ref{tab:starandzsl}. But note that, as a demonstration in \cite{li2017zero}, the VGG19 and ResNet101 features usually achieve better performances than the GoogLeNet features in the ZSL task. Since we use GoogleNet features which are not stronger features, it is fair for such comparisons in Table \ref{tab:starandzsl}.

\begin{table*}[!htb]
		\centering
		\normalsize
		\caption{The standard ZSL classification accuracy (\%). For ImNet, hit @5 is reported. For visual Features: G - GoogleNet\cite{szegedy2015going}; V - VGG19\cite{simonyan2014very}; R - ResNet101\cite{he2016deep}.}
		\begin{tabular}{c |c |c| c|c||c| c|c}
		    \hline
		    \hline
		    \multicolumn{5}{c||}{Medium Datasets} & \multicolumn{3}{c}{Large Dataset}\\
		    \hline
		{Method}  & {V-Features}& AwA & CUB &SUN  &Method  &V-Features & ImNet\\
			\hline
			RPL \cite{shigeto2015ridge} &  G  & 80.4 & 52.4  &- & DeViSE \cite{frome2013devise} & G & 12.8\\
			SSE \cite{zhang2015zero} &  V &  76.3 &  30.4 &- & ConSE\cite{norouzi2013zero} & G & 15.5 \\
			SJE \cite{akata2015evaluation}&  G& 73.9 &  51.7  & 56.1  & AMP\cite{fu2015zero} & G & 13.1\\
			JLSE \cite{zhang2016zerolatent}&  V& 80.5 &  42.1  & - & SS-Voc\cite{fu2016semi}& G & 16.8\\
			SynC \cite{changpinyo2016synthesized}&  G& 72.9 &  54.7  & 62.7  & SAE\cite{kodirov2017semantic} & G & 27.2\\
			SAE \cite{kodirov2017semantic}&  G& 84.7 &  61.4  & 65.2  & CVA\cite{mishra2018generative} & V/R & 24.7\\
			LAD \cite{jiang2017learning}&  V& 82.5 &  56.6  & - & VSZL \cite{wang2018zero} &V &23.1\\
			SCoRe \cite{morgado2017semantically}&  V& 73.9 &  51.7  & - & LESAE
		\cite{liu2018zero} & G & \textbf{27.6}\\
			LESD \cite{ding2017low}&  V/G& 82.8 &  56.2  & - & & &\\
			CVA \cite{mishra2018generative}&  V/R& 71.4 &  52.1  & 61.7  & & &\\
	    	f-CLSWGAN \cite{xian2018feature}&  R& 69.9 &  61.5 & 64.5 & & &\\
			VSZL \cite{wang2018zero}&  V& 85.3 &  57.4  & -  & & &\\
			LESAE \cite{liu2018zero}&  R& 66.1  &  53.9  & 60.0 && & \\
			AAW \cite{kolouri2018joint}&  V& 83.3 &  38.4  & - & & &\\
			LSD\cite{chen2019learning}&  V& 82.7 &  58.5  & - & & &\\
			BZSL\cite{shen2019scalable} & V/G & 80.5 & 56.3 &- & & &\\
			\hline
			JCMSPL ($V\to S$)&  G & { 77.5} &  {60.0 }& \textbf{66.3} & JCMSPL ($V\to S$)&  G &{ 26.0}\\
			JCMSPL ($S\to V$)&  G & \textbf{ 86.2} &  \textbf{62.6 }& {54.1} & JCMSPL ($S\to V$)&  G &{ 27.5}\\
			\hline
			\hline
		\end{tabular}%
		\label{tab:starandzsl}%
	\end{table*}%

\subsubsection{Parameter Settings}
In JCMSPL, there are four tuning parameters $\lambda_1,\dots,\lambda_4$ in Eq . (\ref{equ:JCMSPL}). Following \cite{shigeto2015ridge,kodirov2017semantic}, the parameters are tuned by class-wise cross-validation of the training set. As SUN dataset has multiple splits, in our experiments, we report the average performance of the same splits that are used in \cite{changpinyo2016synthesized}.

\subsubsection{ZSL Settings}
\noindent\textbf{Standard ZSL:} The standard ZSL setting is widely used in recent ZSL works \cite{akata2015evaluation,reed2016learning}. The seen and unseen classes are split following Table \ref{tab:dataset}. \\

\noindent\textbf{Generalized ZSL:} The generalized ZSL setting has recently emerged \cite{rahman2018unified,bucher2017generating}, whose testing set includes both seen and unseen samples. Such setting is clearly more reflective of real world scenarios.

\subsubsection{Evaluation Metrics}
\noindent\textbf{Standard ZSL:} The multi-way classification accuracy as previous works are used for three medium-scale datasets, when the flat hit @K classification accuracy as in \cite{fu2016semi} is used for the large-scale dataset. Hit @K means that for a testing sample, the top K assigned labels should include the correct label. In the experiment, hit @5 accuracy is reported for over all test samples.\\

\noindent\textbf{Generalized ZSL:} Three metrics are used in the generalized ZSL. The first one is $acc_s$, the accuracy of classifying seen samples within all classes, which includes both seen and unseen samples. The second one is, $acc_u$ the accuracy of classifying unseen samples within all classes. The third one is $HM$, which is the harmonic mean of $acc_s$ and $acc_u$, i.e.,
$$HM=\frac{2 \times acc_s \times acc_u}{acc_s+acc_u}.$$

\subsubsection{Competitive Methods}
16 existing ZSL methods are used for performance comparison for three medium-scale datasets and 8 state-of-art are used for the large-scale ones, where RPL \cite{shigeto2015ridge}, SSE \cite{zhang2015zero}, SJE \cite{akata2015evaluation}, JLSE \cite{zhang2016zerolatent}, SynC \cite{changpinyo2016synthesized}, SAE \cite{kodirov2017semantic}, LAD \cite{jiang2017learning}, SCoRe \cite{morgado2017semantically}, LESD \cite{ding2017low}, CVA \cite{mishra2018generative}, f-CLSWGAN \cite{xian2018feature}, VSZL \cite{wang2018zero}, LESAE \cite{liu2018zero}, AAW \cite{kolouri2018joint}, LSD \cite{chen2019learning} and BZSL\cite{shen2019scalable} are used for medium-scale ones, and DeViSE \cite{frome2013devise}, ConSE \cite{norouzi2013zero}, AMP \cite{fu2015zero}, SS-Voc \cite{fu2016semi}, SAE \cite{kodirov2017semantic},  CVA \cite{mishra2018generative}, VSZL \cite{wang2018zero}, and LESAE \cite{liu2018zero} are used for the large one. These ZSL methods cover a wide range of recent and representative ZSL models and achieve state-of-the-art results.

\subsection{Experiment Results}
\subsubsection{Standard ZSL}
The comparative result of various datasets under standard ZSL settings are listed in Table \ref{tab:starandzsl}. All these comparative results are based on inductive ZSL. That is, no unlabeled unseen samples are incorporated in the training phase. Based on Table \ref{tab:starandzsl}, our method achieves the best results on all three medium-scale datasets and a comparable result on the large dataset, which demonstrates that the reconstructions of both visual and semantic mitigate the domain shift problem and the class-specific distinct common space makes the visual and semantic features match more precisely. For all three medium datasets, our method improves about 1\% performance over the strongest competitors. For the large dataset, although our method is only slightly lower than the LESAE, it is still much better than other competitive methods.

\subsubsection{Generalized ZSL}

\begin{table}[!htb]
		\centering
		\normalsize
		\caption{The comparative results(\%) of generalized ZSL as setting in \cite{chao2016empirical}. All the results are tests on GoogleNet visual features.}
		\begin{tabular}{c |c c c| c c c}
			\hline
			\hline
			\multirow{2}{*}{Method} & \multicolumn{3}{|c|}{AwA} &\multicolumn{3}{c}{CUB} \\ \cline{2-7}
			 & $acc_s$ & $acc_u$ & {$HM$}& $acc_s$ & $acc_u$ & {$HM$}\\
			\hline
			DAP \cite{lampert2014attribute} &  77.9 & 2.4 & 4.7 & 55.1 &4.0 &7.5\\
			IAP \cite{lampert2014attribute} &  76.8 & 1.7 & 3.3 & 69.4 & 1.0 & 2.0\\
			ConSE \cite{norouzi2013zero} &  75.9 & 9.5 & 16.9 & 69.9 &1.8 &2.5\\
			APD \cite{rahman2018unified} &  43.2 & \textbf{61.7} & 50.8 & 23.4 & 39.9 & 29.5\\
			GAN \cite{bucher2017generating} & \textbf{81.3} & 32.3 & 46.2 & \textbf{72.0} & 26.9 & 39.2\\
			SAE \cite{kodirov2017semantic}& 67.6 & 43.3 & \textbf{52.8} & 36.1 &28.0 &31.5\\
			\hline
			JCMSPL (ours) &  48.3  & 56.4 &  52.1 & 54.2 & \textbf{50.7} & \textbf{52.4}\\
			\hline
			\hline
		\end{tabular}%
		\label{tab:generalizedzsl}%
	\end{table}%

Following the same setting of \cite{chao2016empirical}, we extract out $20\%$ of seen class data samples and mix them with the unseen class samples. The generalized ZSL of AwA and CUB are listed in Table \ref{tab:generalizedzsl}, which includes 6 competitive methods. Although the $HM$ score of AwA is slightly lower than SAE, it is still comparable, and its $acc_u$, the unseen to all class classification accuracy, is still higher than SAE, thus demonstrating a better generalization capability. For the CUB dataset, our method achieves the highest $HM$ and $acc_u$, which again shows that our method is favored over the generalized ZSL setting. The high accuracy for $acc_s$ reflects that the method is overfitting when training for seen classes and is difficult to generalize to unseen ones.


\subsection{Further Evaluations}

\subsubsection{Ablation Study}
\begin{figure}[htb]
		\centering
		\includegraphics[width=0.45\textwidth]{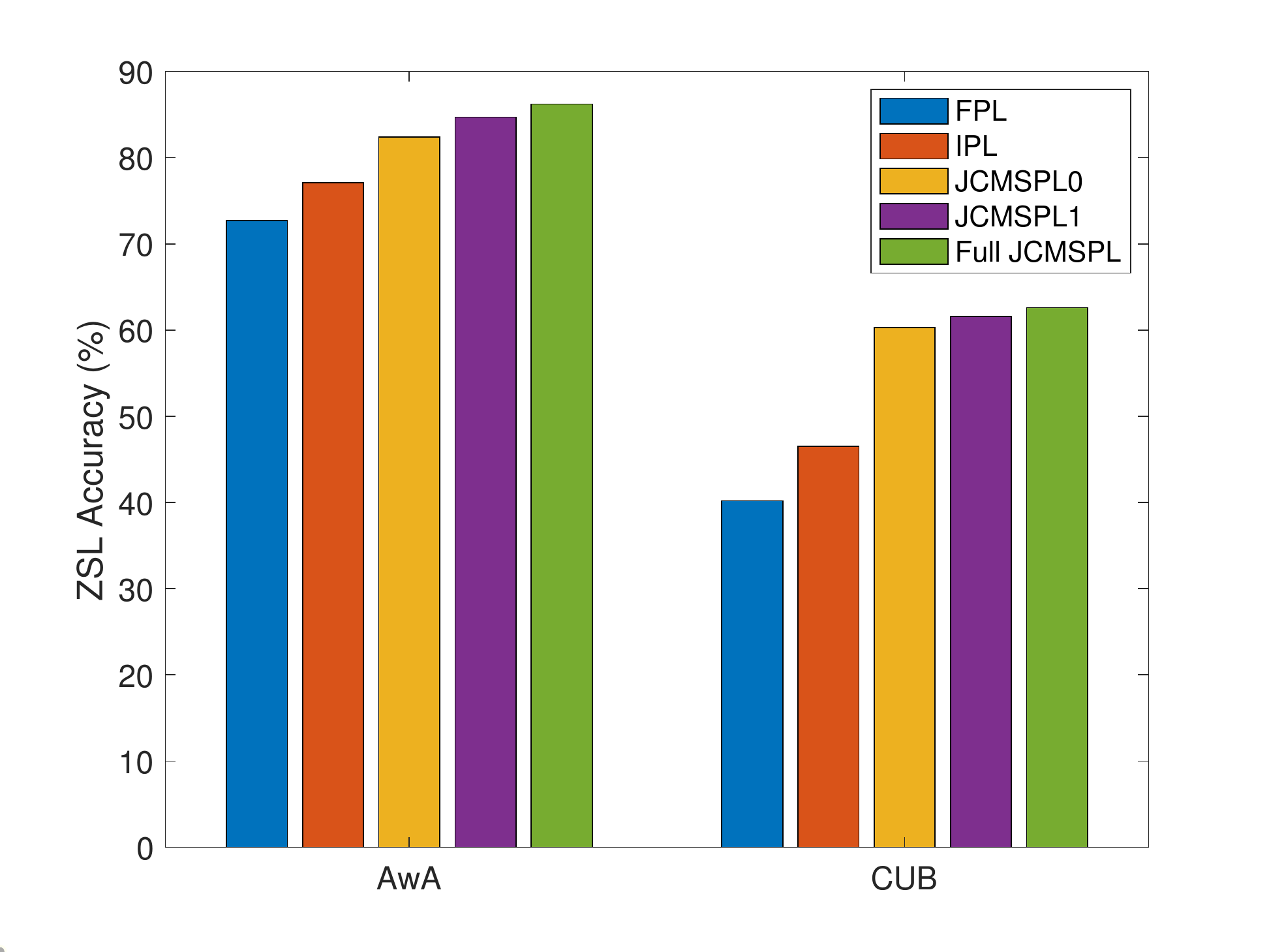}
		\caption{Ablation study results of standard ZSL on two medium-scale datasets.}
		\label{fig:simpleJCMSPL}
\end{figure}

Our JCMSPL model can also be simplified as follows:
\begin{itemize}
    \item [(1)] When $\lambda_2=0$, then the class-specific information is not used, our JCMSPL reduces to the joint space projection with reconstruction, and denoted as JCMSPL1, \textit{i.e.,}
        \begin{equation*}
        \begin{split}
        \underset{\mathbf{A},\mathbf{B},\mathbf{C}}{\min}~
    &\frac{1}{2}\|\mathbf{A}\mathbf{X_s}-\mathbf{C}\|_F^2+\frac{\lambda_1}{2}\|\mathbf{B}\mathbf{Y_s}-\mathbf{C}\|_F^2\\
    &+\frac{\lambda_3}{2}\|\mathbf{X_s}-\mathbf{A^T}\mathbf{C}\|_F^2+\frac{\lambda_4}{2}\|\mathbf{Y_s}-\mathbf{B^T}\mathbf{C}\|_F^2.
        \end{split}
    \end{equation*}
    \item [(2)] When $\lambda_3=0$ and $\lambda_4=0$, the class-specific information is used, but the reconstructions of both visual and semantic space are not taken in account. Our JCMSPL reduces to the joint concept matching space projection without reconstruction, which is denoted as JCMSPL0, \textit{i.e.,}
    \begin{equation*}
        \begin{split}
        \underset{\mathbf{A},\mathbf{B},\mathbf{C}}{\min}~
    &\frac{1}{2}\|\mathbf{A}\mathbf{X_s}-\mathbf{C}\|_F^2+\frac{\lambda_1}{2}\|\mathbf{B}\mathbf{Y_s}-\mathbf{C}\|_F^2\\
    &+\frac{\lambda_2}{2} \|\mathbf{C}-\mathbf{H}\|_F^2.\\
        \end{split}
    \end{equation*}
    This is similar to the third group of projection learning in the literature but with the class information.
    \item [(3)] When $\lambda_2=0, \lambda_3=0$ and $\lambda_4=0$, both the class-specific information and the reconstructions of both visual and semantic space are not used. Our JCMSPL then reduces to the joint concept matching space projection without reconstruction and any class-specific information, which is similar to the third group of projection learning in the literature \cite{changpinyo2016synthesized,zhang2016zerolatent}. This is denoted as intermediate space projection learning (IPL), \textit{i.e.,}
    \begin{equation*}
        \begin{split}
        \underset{\mathbf{A},\mathbf{B},\mathbf{C}}{\min}~
    &\frac{1}{2}\|\mathbf{A}\mathbf{X_s}-\mathbf{C}\|_F^2+\frac{\lambda_1}{2}\|\mathbf{B}\mathbf{Y_s}-\mathbf{C}\|_F^2.\\
        \end{split}
    \end{equation*}
    \item [(4)] When $\lambda_1,\lambda_2, \lambda_3$ and $\lambda_4=0$ and using the semantic space instead of the intermediate/common space, JCMSPL is finally reduced to the original forward projection learning method \cite{akata2015evaluation}, denoted as FPL,
    \textit{i.e.,}
    \begin{equation*}
        \begin{split}
        \underset{\mathbf{A}}{\min}~
    &\frac{1}{2}\|\mathbf{A}\mathbf{X_s}-\mathbf{Y_s}\|_F^2.
        \end{split}
    \end{equation*}
\end{itemize}
To evaluate the contribution of proposed Full JCMSPL method, its simple reduced FPL, IPL, JCMSPL0, JCMSPL1 are compared with same standard splits of AwA and CUB datasets. The standard ZSL accuracy of hereabove simple JCMSPL methods are listed and shown in Table \ref{tab:simpleJCMSPL}
and Fig. \ref{fig:simpleJCMSPL}.

\begin{table}[!htb]
		\centering
		\normalsize
		\caption{The evaluation of the contributions of JCMSPL and the improtance of both reconstruction constraint and class-specific information.}
		\begin{tabular}{l l l}
		\hline
			\hline
		    Projection Method & AwA & CUB \\
			\hline
			FPL & 72.7\%  & 40.2\%  \\
			IPL &  77.1\% & 46.5\%  \\
			JCMSPL0&  82.4\% & 60.3\%  \\
			JCMSPL1 & 84.7\% & 61.6\% \\
			\hline
			Full JCMSPL & \textbf{86.2}\% &  \textbf{62.6}\% \\
			\hline
			\hline
		\end{tabular}%
		\label{tab:simpleJCMSPL}%
	\end{table}%

\begin{figure*}[!htb]
	\centering
    \subfloat[$\lambda_1$]{\includegraphics[width=0.43\textwidth]{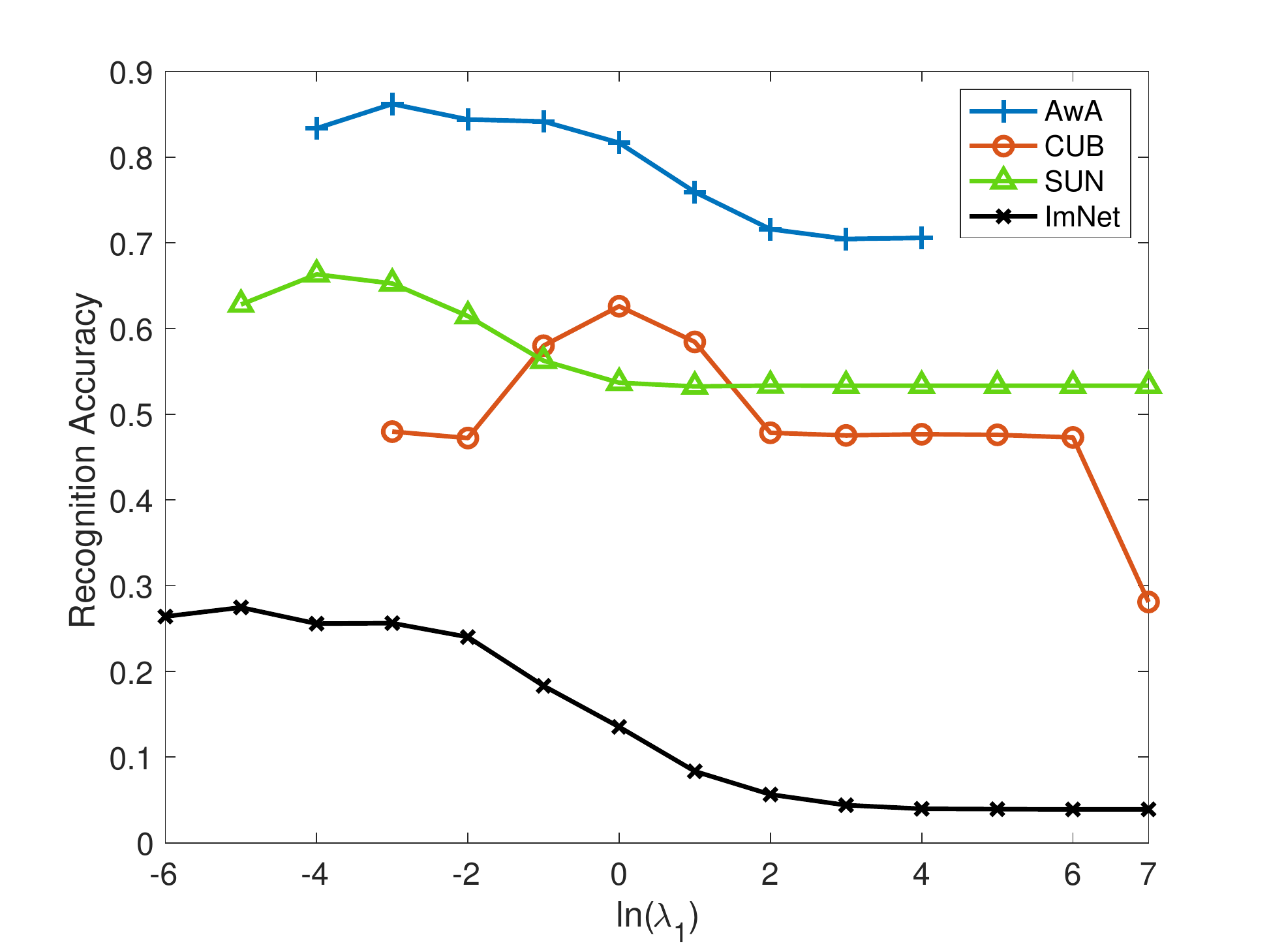}\label{fig:f1}\label{fig:paramlambda1}}
	\quad
	\subfloat[$\lambda_2$]{\includegraphics[width=0.43\textwidth]{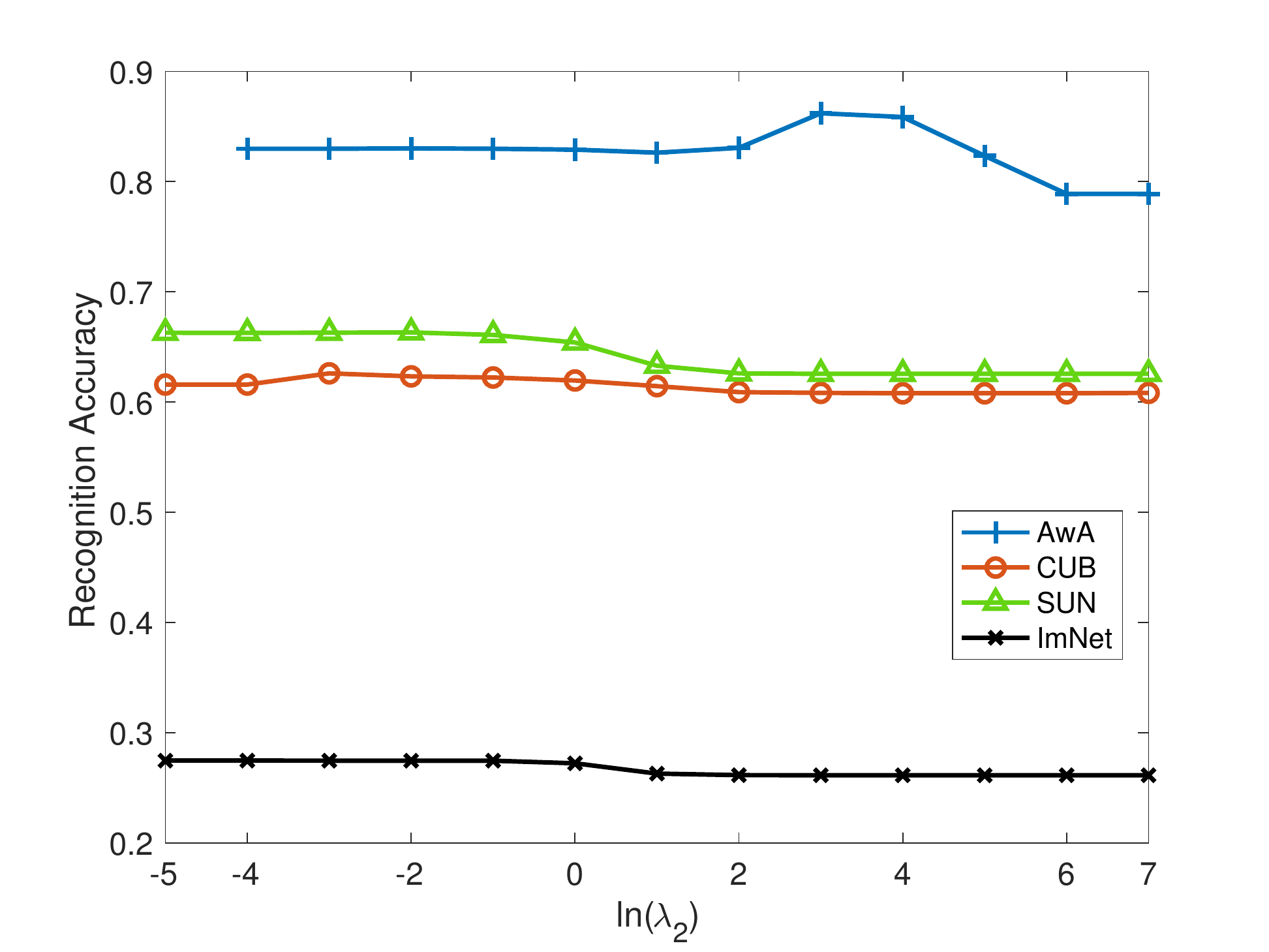}\label{fig:param2}\label{fig:paramlambda2}}
	\hfill
	\subfloat[$\lambda_3$]{\includegraphics[width=0.43\textwidth]{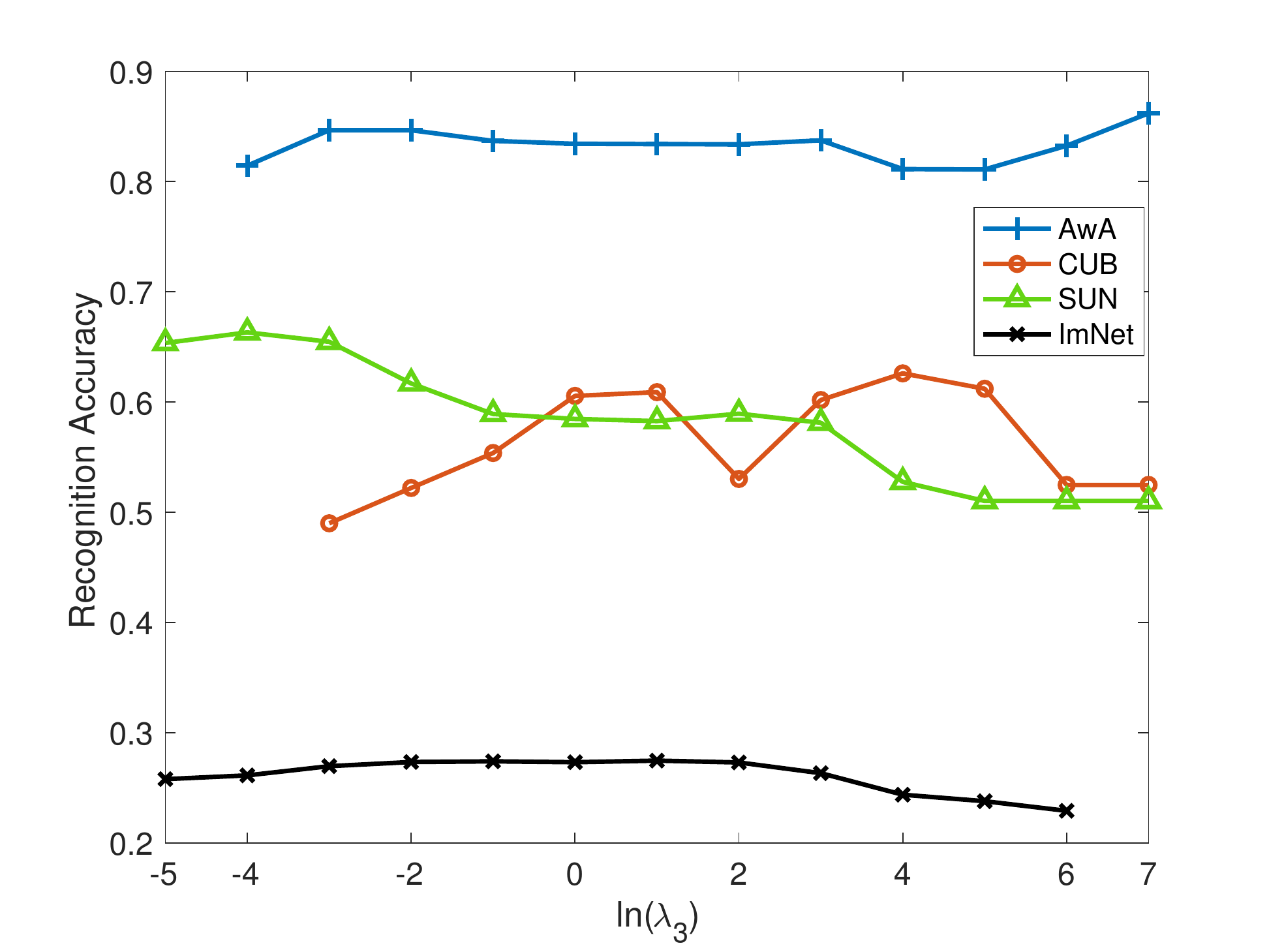}\label{fig:param3}\label{fig:paramlambda3}}
	\quad
	\subfloat[$\lambda_4$]{\includegraphics[width=0.43\textwidth]{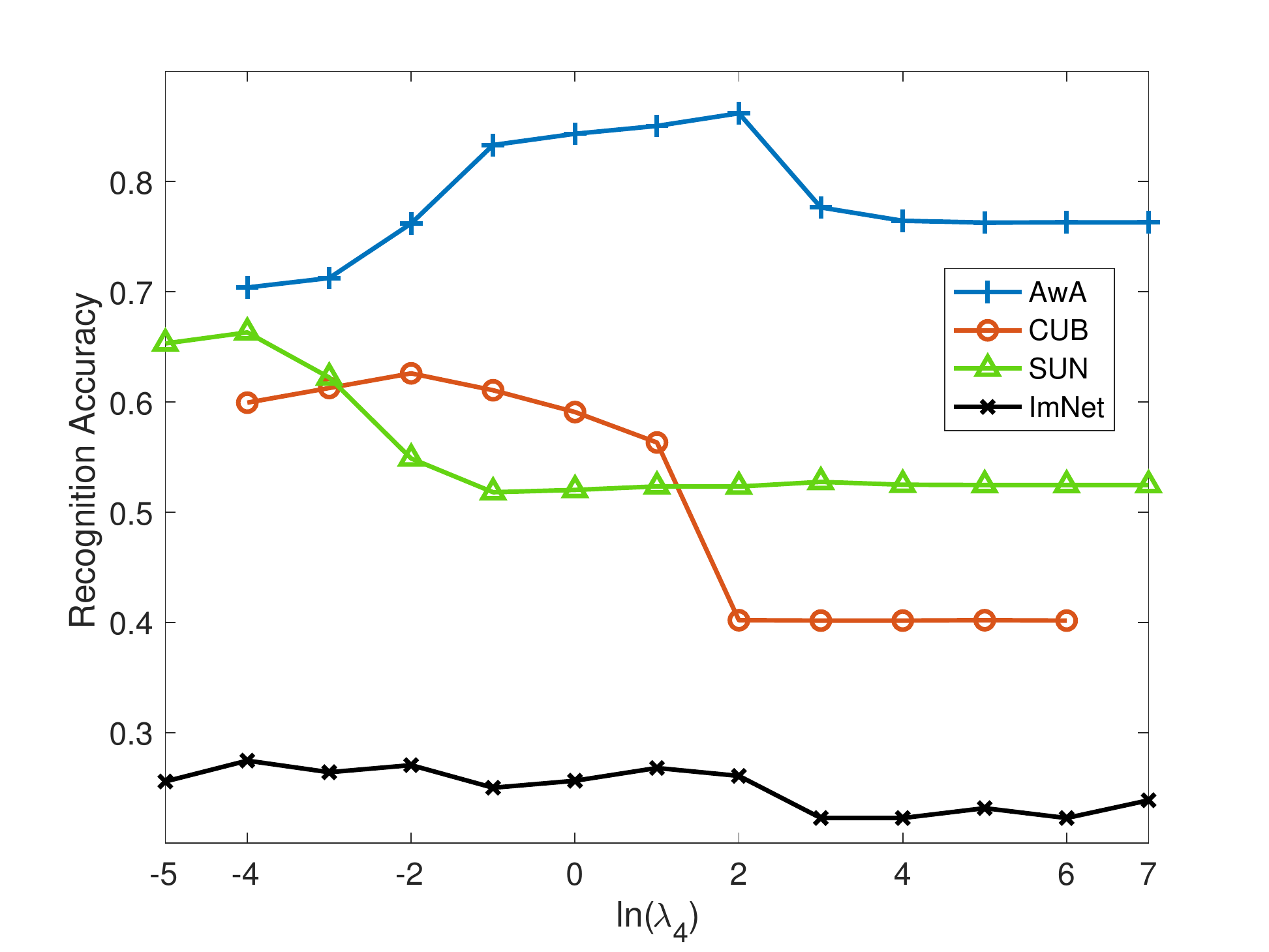}\label{fig:param4}\label{fig:paramlambda4}}
	\caption{Parameters analysis for $\lambda_1,\dots,\lambda_4$.}
	\label{fig:paramanalysis}
\end{figure*}

The ablation study results shows that: {(1) {When comparing the IPL and FPL, it shows that the intermediate space respectively brings $4.4\%$ and $6.3\%$ improvements for AwA and CUB datasets. The common space introduces the latent representations of both visual and semantic features that enhance their similarities.} (2) The result of JCMSPL0 = IPL + class-specific information also has a $5\%-13\%$ gain in comparison to IPL, which validates the effectiveness of class-specific information. Even win presence of good results, the class-specific information also introduce a great than $1\%$ gains, as comparing JCMSPL1 with Full JCMSPL. Class-specific latent space well matches the visual and semantic features class-to-class. (3) As the shown result of JCMSPL1 = IPL + reconstructions vs. IPL, both reconstructions of visual and semantic features are important for ZSL, which results in $7\%-14\%$ improvements. (4) To combine with all these class-specific information and feature reconstructions, our full JCMSPL has significant improvements ranging from $9\%-16\%$ when comparing to IPL and achieves $1\%-2\%$ improvements when comparing to JCMSPL1, the one only with such reconstruction constraints. }

\subsubsection{Parameters Analysis}

Figure \ref{fig:paramanalysis} shows that the values of $\lambda_1,\dots,\lambda_4$ that achieve the best performance in different datasets. The detailed settings of four benchmark datasets are listed in Table \ref{tab:paramanalysis}.

\begin{table}[htb]
		\centering
		\normalsize
		\caption{The parameters $\lambda_1,\dots,\lambda_4$ settings are used for four different datasets.}
		\begin{tabular}{l l l l l}
		\hline
			\hline
		    Parameter & AwA & CUB & SUN & ImNet\\
			\hline
			$\lambda_1$ & $1e-3$  & $1$ &$1e-4$&$1e-5$\\
			$\lambda_2$ & $1e3$ & $1e-3$&$1e-2$&$1e-5$ \\
			$\lambda_3$&  $1e7$ & $1e4$ &$1e-4$& $1e1$\\
			$\lambda_4$ & $1e2$& $1e-1$ & $1e-4$ & $1e-4$\\
			\hline
			\hline
		\end{tabular}%
		\label{tab:paramanalysis}
	\end{table}%

First we evaluate the effect of the constraint that imposed by projecting the semantic features to common distinct space by tuning the $\lambda_1$ during our algorithm training phase. Since visual features have different structures from semantic features, the weights of both visual feature mapping and semantic features mapping should be different, too. In Figure \ref{fig:paramlambda1}, it is shown that our algorithm achieves better performance, when the weights of semantic feature mapping is in the range of $[10^{-5},10]$ for all four datasets.

We then evaluate $\lambda_2$, the influence of class-specific information term during training. From the evaluation in Figure \ref{fig:paramlambda2}, it is readily seen that the classification accuracy tends to be better during the range of $[10^2,10^6]$ for AwA, the range of $[10^{-3},10^{-2}]$ for CUB, the range of $[10^{-5},10^{-2}]$ for SUN and the range of $[10^{-5},1]$ for ImNet. Comparing with the magnitude of other parameters in each dataset, the class-specific information term has large weights for AwA and SUN datasets, which imparts the distinct knowledge of different class for a better matching between visual and semantic features. Although its weight is small for CUB dataset, the accuracy drops when its weight $\lambda_2$ decreases. This validates that the class-specific information term is also helpful for zero-shot recognition to be robust against the domain shift issue, and match the same class visual and semantic features.

In addition, through the analysis of parameter $\lambda_3$ and $\lambda_4$ (Figure \ref{fig:paramlambda3} and \ref{fig:paramlambda3} ), they show that the ranges of both $\lambda_3$ and $\lambda_4$ that achieve promising performances, are respectively less than $10^8$ and $10^3$ on four datasets. Although four different datasets have different magnitudes, the $\lambda_3$ and $\lambda_4$ are relatively larger than $\lambda_1$ and $\lambda_2$ (the detailed values is listed in Table \ref{tab:paramanalysis}). This shows that the reconstructions of both visual and semantic features can improve the zero-shot learning ability, narrow the domain shift gap, and explore more intrinsic structure within seen data.

\subsubsection{Convergence and Complexity Analysis}
\noindent \textbf{Convergence Analysis:} In light of the non-convexity of Eq. (\ref{equ:JCMSPL}), the convergence of our algorithm is not guaranteed by standard results. We hence separately prove the convergence of our algorithm: let $f(\mathbf{A},\mathbf{B},\mathbf{C})$ be the loss function of Eq. (\ref{equ:JCMSPL}), then the following result follows.

	\begin{theorem}\label{theorem:main}
		The sequence $\{\Theta_t=(\mathbf{A}_t,\mathbf{B}_t,\mathbf{C}_t)\}_{t=1}^\infty$ converges to the following set of feasible stationary points of the loss function $f$, which is bounded by a universal constant $R$ depending on the initialization \footnote{The norm $\|\Theta\|$ is any norm which is continuous with respect to the 2-norm of the components, for example their sum of 2-norms.}:
		\[Q=\{\Theta=(\mathbf{A},\mathbf{B},\mathbf{C})\mid\|\Theta\|<R\}.\]
	\end{theorem}
Theorem \ref{theorem:main} shows that Algorithm 1 not only converges, but also generates a solution sequence that eventually converges to the stationary points of the underlying optimization. Theorem \ref{theorem:main} has been proved in the Appendix \ref{sec:proof}.
\\

\noindent \textbf{Complexity Analysis:} Since the solution complexity of sylvester equation only depends on the dimension of rows, the complexities of updating $\mathbf{A}$ and $\mathbf{B}$ only related with the dimension of features, but instead of the sample size $n$. That is, the complexities of updating $\mathbf{A}$ and $\mathbf{B}$ are respectively $O(m^3)$ and $O(d^3)$, while the complexity of updating $\mathbf{C}$ is $O(k^3+k^2m+k^2d+kmn_s+kdn_s)\leq O(max(k,m,d,n_s)^3)$.

\begin{figure}[!htb]
		\centering
		\includegraphics[width=0.45\textwidth]{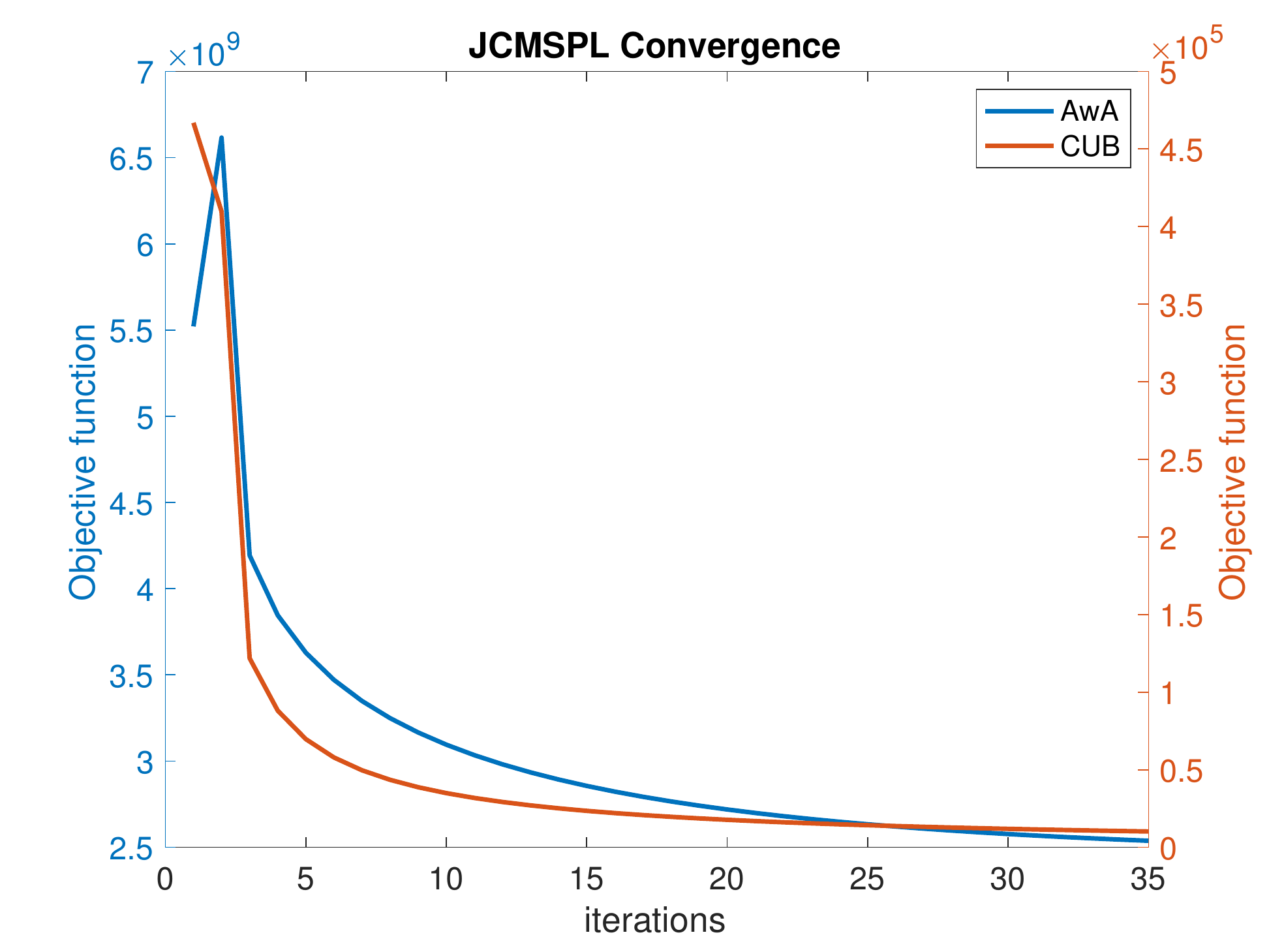}
		\caption{Convergence of JCMSPL on AwA and CUB Datasets. The left-side vertical axis is for AwA dataset, while the right-side vertical axis is for CUB dataset. }
		\label{fig:converge}
\end{figure}

The empirical results also show that our proposed JCMSPL algorithm converges very fast. Figure \ref{fig:converge} illustrates the converges curve of JCMSPL on two medium-scale datasets. It is clearly shown that the value of our objective function decreases quickly and stabilize at $35$, and our algorithm converges in 35 iterations for all four datasets. Consequently, both of convergence and complexity analysis and empirical results all demonstrate that our proposed algorithm JCMSPL is practical to realistic problems by its benefits of good convergence property and low complexity.

\section{Conclusion}\label{sec:conclusion}
We purposed a novel inductive ZSL method by incorporating the class-specific information in a common latent space and the reconstructions of both visual and semantic features. In contrast to most of the existing ZSL methods, they neither consider the reconstructions of features nor involve the class-specific information in latent common space. Such class-specific latent space provides more distinct information, and such reconstructions also enhance the robustness by mitigating the domain shifts. Our proposed JCMSPL leverages the intrinsic structure of visual and semantic features as well as their class-level matching. An efficient algorithm is developed and followed by a theoretically rigorous algorithm analysis. Extensive experiments on four benchmark datasets demonstrate that our proposed JCMSPL method yields superior classification performances for both standard and generalized ZSL than other well-established inductive ZSL methods.


%

\appendices
\section{Proof Convergence of Algorithm 1} \label{sec:proof}
From the Algorithm 1 in our paper, our primal problem can be written as follows:
\begin{equation}
\label{equ:JCMSPLpriaml}
\begin{split}
    &f(\mathbf{A},\mathbf{B},\mathbf{C})=~ \frac{1}{2}\|\mathbf{A}\mathbf{X_s}-\mathbf{C}\|_F^2+\frac{\lambda_1}{2}\|\mathbf{B}\mathbf{Y_s}-\mathbf{C}\|_F^2\\
    &+\frac{\lambda_2}{2}\|\mathbf{C}-\mathbf{H}\|_F^2+\frac{\lambda_3}{2}\|\mathbf{X_s}-\mathbf{A^T}\mathbf{C}\|_F^2+\frac{\lambda_4}{2}\|\mathbf{Y_s}-\mathbf{B^T}\mathbf{C}\|_F^2,
\end{split}
\end{equation}
And, our algorithm in the paper can also be written as Algorithm \ref{alg:learning_JCMSPL}.

\begin{algorithm}[htb]
	\caption{Joint Concept Matching-Space Projection Learning}\label{alg:learning_JCMSPL}
		At each iteration $t+1$, compute:
		 \begin{equation}\label{equ:updateA_p}
		 \mathbf{A}_{t+1}=\arg\min_\mathbf{A} f(\mathbf{A}_t,\mathbf{B}_t,\mathbf{C}_t);
		 \end{equation}
		 \begin{equation}\label{equ:updateB_p}
		 \mathbf{B}_{t+1}=\arg\min_\mathbf{B} f(\mathbf{A}_{t+1},\mathbf{B}_t,\mathbf{C}_t);
		 \end{equation}
		 \begin{equation}\label{equ:updateC_p}
		 \mathbf{C}_{t+1}=\arg\min_\mathbf{C} f(\mathbf{A}_{t+1},\mathbf{B}_{t+1},\mathbf{C}_t);
		 \end{equation}
\end{algorithm}

We take $f_t=f(\mathbf{A}_t,\mathbf{B}_t,\mathbf{C}_t)$ for $t=0,1,2,\ldots$ and note that the change in $f_t$ can be controlled by the following result:
\begin{theorem}\label{theorem:iter}
	\begin{equation}\label{equ:iter}
	\begin{split}
	f_{t+1}-f_t &\leq -\frac{m^t_A}{2}\|\mathbf{A}_{t+1}-\mathbf{A}_t\|^2-\frac{m^t_B}{2}\|\mathbf{B}_{t+1}-\mathbf{B}_t\|^2\\
	&-\frac{m^t_C}{2}\|\mathbf{C}_{t+1}-\mathbf{C}_t\|^2,\\
	\end{split}
	\end{equation}
	where
	\[
	m^t_A=\sigma_{min}(\lambda_3\mathbf{C}_t\mathbf{^T}\mathbf{C}_t+ \mathbf{X^T_s}\mathbf{X_s})\geq \sigma_{min}(\mathbf{X^T_s}\mathbf{X_s})>0,
	\]
	\[
	m^t_B=\sigma_{min}(\lambda_4\mathbf{C}_t\mathbf{^T}\mathbf{C}_t+ \lambda_1\mathbf{Y^T_s}\mathbf{Y_s})\geq \sigma_{min}(\lambda_1\mathbf{Y^T_s}\mathbf{Y_s})>0,
	\]
	\[
	m^t_C=\sigma_{min}((1+\lambda_1+\lambda_2)\mathbf{I}+\lambda_3 \mathbf{A_{t+1}A^T_{t+1}}+\lambda_3 \mathbf{B_{t+1}B^T_{t+1}})\geq 1.
	\]
	The sequence $\{f_t\}_{t=1}^\infty$ is positive and decreasing, hence convergent.
	\begin{proof}
		Respectively denote by $\Delta f_{t,A},\Delta f_{t,B},\Delta f_{t,C}$, the change in $f$ corresponding to the update of $\mathbf{A},~\mathbf{B},~\mathbf{C}$ in Eq. (\ref{equ:updateA_p}) - (\ref{equ:updateC_p}). Notice that
		\[
		f_{t+1}-f_t=\Delta f_{t,A}+\Delta f_{t,B}+\Delta f_{t,C}.
		\]
		The function $g(\mathbf{A})=f(\mathbf{A},\mathbf{B}_{t},\mathbf{C}_t)$ is quadratic and $m_A$ is strongly convex, where $m_A$ is the smallest singular value of Hessian. Hence,
		\begin{equation}\label{eq:boundA}
		\Delta f_{t,A}=g(\mathbf{A}_{t+1})-\min_\mathbf{A} g(\mathbf{A})\leq-\frac{m^t_A}{2}\|\mathbf{A}_{t+1}-\mathbf{A}_t\|^2.
		\end{equation}
		Similarly, for updating $\mathbf{B}$ and $\mathbf{C}$ , we obtain
		\begin{equation}\label{eq:boundB}
		\Delta f_{t,B}=g(\mathbf{B}_{t+1})-\min_\mathbf{B} g(\mathbf{B})\leq-\frac{m^t_B}{2}\|\mathbf{B}_{t+1}-\mathbf{B}_t\|^2,
		\end{equation}
		\begin{equation}\label{eq:boundC}
		\Delta f_{t,C}=g(\mathbf{C}_{t+1})-\min_\mathbf{C} g(\mathbf{C})\leq-\frac{m^t_C}{2}\|\mathbf{C}_{t+1}-\mathbf{C}_t\|^2.
		\end{equation}
		Summing the inequalities in Eq.~\eqref{eq:boundA}, Eq.~\eqref{eq:boundB}, and Eq.~\eqref{eq:boundC} then Eq. (\ref{equ:iter}) is proved.

		Our primal lost function is as follow
		\begin{equation}\label{eq:Le}
		    \begin{split}
		    &f_{t,e}(\mathbf{A},\mathbf{B},\mathbf{C})=\frac{1}{2}\|\mathbf{A}\mathbf{X_s}-\mathbf{C}\|_F^2+\frac{\lambda_1}{2}\|\mathbf{B}\mathbf{Y_s}-\mathbf{C}\|_F^2\\
    &+\frac{\lambda_2}{2}\|\mathbf{C}-\mathbf{H}\|_F^2+\frac{\lambda_3}{2}\|\mathbf{X_s}-\mathbf{A^T}\mathbf{C}\|_F^2+\frac{\lambda_4}{2}\|\mathbf{Y_s}-\mathbf{B^T}\mathbf{C}\|_F^2,
		    \end{split}
		\end{equation}
		Hence, we have $f_{t,e} \geq 0$. In particular, we obtain $ f_{t+1}=f_{t,e}(\mathbf{A}_{t+1},\mathbf{B}_{t+1},\mathbf{C}_{t+1})\geq 0$. Now, we use complete (strong) induction to show that $f_{t+1} \leq f_t$ for t = 1,2,.... Suppose that this holds for t = 1,2,...,k. We conclude that $f_t \leq f_1$. Since $m^t_A>0$, $m^t_B>0$ and $m^t_C>1, \forall t$, according to Eq. (\ref{equ:iter}), we conclude that $f_{t+1}\leq f_t \leq f_1$ which completes the proof.
	\end{proof}
\end{theorem}

We finally obtain the following corollary which clarifies the statement and gives the proof of our main result in Theorem \ref{theorem:main}:
\newtheorem{cor}{Corollary}
\begin{cor}
	The sequence $\{\Theta_t=(\mathbf{A}_t,\mathbf{B}_t,\mathbf{C}_t)\}_{t=1}^\infty$ satisfies the following:
	\begin{enumerate}
		\item[a.] The parameters for $t=0,1,2,...$ are bounded by $R$ which only depends on the initialization, i.e \[\|\Theta_t\|=\max\left\{\|\mathbf{A}_t\|,\|\mathbf{B}_t\|,\|\mathbf{C}_t\|\right\}<R.\] Hence, the are confined in a compact set.
		\item[b.] Any convergence subsequence of $\{\Theta_t\}$ converges to a point $\Theta^*\in Q$.
		\item[c.] $\mathrm{dist}(\Theta_t,Q)$ converges to zero, where
		\[
		\mathrm{dist}(\Theta,Q)=\min_{\Theta^\prime\in Q}\|\Theta^\prime-\Theta\|。
		\]
	\end{enumerate}
	\begin{proof}
		Part a is simply obtained by noticing \eqref{eq:Le} and the fact that $f_{t,e}(\mathbf{A}_t,\mathbf{B}_t,\mathbf{C}_t)=f_t\leq f_1$, since $\{f_t\}$ is decreasing. For part b, note that since the sequence $\{f_t\}$ is convergent, we have $\lim_{t\to\infty}f_{t+1}-f_t=0$, which according to \eqref{equ:iter} yields
		\[
		\lim_{t\to\infty}\|\mathbf{A}_{t+1}-\mathbf{A}_{t}\|_2^2=\lim_{t\to\infty}\|\mathbf{B}_{t+1}-\mathbf{B}_{t}\|_2^2
		=\lim_{t\to\infty}\|\mathbf{C}_{t+1}-\mathbf{C}_{t}\|_2^2.
		\]
		Moreover, note that the loss function $f$ is $f_A-$second order Lipschitz with respect to $\bf{A}$ (fixing the rest) with $f_A=\|\lambda_3\mathbf{C^T_{t}}\mathbf{C}_{t}+\mathbf{X^T_{s}}\mathbf{X_{s}}\|_*$. We obtain that
		\[
		\left\|
		\nabla_A
		f(\mathbf{A}_t,\mathbf{B}_t,\mathbf{C}_t)\right\|_2^2\leq f_A^2\|\mathbf{A}_{t+1}-\mathbf{A}_{t}\|_2^2，
		\]
		which yields
		\[
		\lim_{t\to\infty}
		\left\|
		\nabla_A
		f(\mathbf{A}_{t},\mathbf{B}_{t},\mathbf{C}_{t})\right\|_2^2，
		=0
		\]
		Similarly, we obtain
		\[
		\lim_{t\to\infty}
		\left\|
		\nabla_{B}
		f(\mathbf{A}_{t+1},\mathbf{B}_{t},\mathbf{C}_{t})\right\|_2^2
		=0，
		\]
		\[
		\lim_{t\to\infty}
		\left\|
		\nabla_{C}
		f(\mathbf{A}_{t+1},\mathbf{B}_{t+1},\mathbf{C}_{t})\right\|_2^2
		=0
		\]
		Now, take a subsequence of $\{\Theta_t\}$ converging to a point $\Theta_*=(\mathbf{A^*},\mathbf{B^*},\mathbf{C^*})$. Since the argument of the above limits are continuous we obtain
		\[
		\nabla_{A}
	f(\Theta_*)=0,\quad \nabla_B
		f(\Theta_*)=0,\quad  \nabla_{C}
		f(\Theta_*)=0
		\]
		Therefore, $\Theta_*\in Q$. For part c, suppose that the claim is not true. Then, according to part a there exists a convergent subsequence of $\{\Theta_t\}$ which is $\epsilon-$distant from $Q$, $i.e.$, $\mathrm{dist}(\Theta_k,Q)=\epsilon>0$. Then, the convergence point is also $\epsilon-$distant from $Q$ which contradicts part b and completes the proof.

	\end{proof}
\end{cor}

\ifCLASSOPTIONcompsoc
  \section*{Acknowledgments}
  We gratefully acknowledge the generous support of the U.S. Army Research Office DURIP under grant W911NF1810209.
\else
  \section*{Acknowledgment}
\fi


\ifCLASSOPTIONcaptionsoff
  \newpage
\fi



%


\bibliographystyle{IEEEtran}
\bibliography{IEEEabrv,egbib}

\end{document}